\def\year{2020}\relax
\documentclass[letterpaper]{article} 
\usepackage{aaai20}  
\usepackage{times}  
\usepackage{helvet} 
\usepackage{courier}  
\usepackage[hyphens]{url}  
\usepackage{graphicx} 
\usepackage{graphicx}  
\usepackage{xcolor}
\usepackage{amsmath}
\usepackage{amsfonts}

\title{Automated Biodesign Engineering by Abductive Meta-Interpretive Learning}
\author{
  \Large \textbf{Wang-Zhou Dai\textsuperscript{\rm 1}, Liam Hallett\textsuperscript{\rm 2}, Stephen H. Muggleton\textsuperscript{\rm 1}, Geoff S. Baldwin\textsuperscript{\rm 2}}\\ 
  \textsuperscript{\rm 1}Department of Computing, Imperial College London, SW7 2AZ, UK. \\
  \textsuperscript{\rm 2}Department of Life Science, Imperial College London, SW7 2AZ, UK.\\ 
  \{w.dai, l.hallett19, s.muggleton, g.baldwin\}@imperial.ac.uk
}
\begin{document}

\maketitle

\begin{abstract}
The application of Artificial Intelligence (AI) to synthetic biology will provide the foundation for the creation of a high throughput automated platform for genetic design, in which a learning machine is used to iteratively optimise the system through a design-build-test-learn (DBTL) cycle. However, mainstream machine learning techniques represented by deep learning lacks the capability to represent relational knowledge and requires prodigious amounts of annotated training data. These drawbacks strongly restrict AI's role in synthetic biology in which experimentation is inherently resource and time intensive. In this work, we propose an automated biodesign engineering framework empowered by Abductive Meta-Interpretive Learning ($Meta_{Abd}$), a novel machine learning approach that combines symbolic and sub-symbolic machine learning, to further enhance the DBTL cycle by enabling the learning machine to 1) exploit domain knowledge and learn human-interpretable models that are expressed by formal languages such as first-order logic; 2) simultaneously optimise the structure and parameters of the models to make accurate numerical predictions; 3) reduce the cost of experiments and effort on data annotation by actively generating hypotheses and examples. To verify the effectiveness of  $Meta_{Abd}$, we have modelled a synthetic dataset for the production of proteins from a three gene operon in a microbial host, which represents a common synthetic biology problem.
\end{abstract}

\section{Introduction}
The development of Synthetic Biology has created a lot of new tools for building genetic designs and the application of automation is now beginning to impact the field, with the possibility of creating large datasets. In parallel, Artificial Intelligence has made huge impacts in its application to the analysis and interpretation of big data, including biological data. However, the true potential of the intersection of the two areas is still far from fulfilled, as it is still unclear how AI can facilitate the Design-Build-Test-Learn (DBTL) cycle. 

Synthetic biology seeks to optimise biological systems for specific functionalities, such as biosynthesis, biosensing, and genetic circuits. The systems implemented in such applications typically consist of genetic devices comprised of discrete genetic parts whose background knowledge and hypotheses are usually symbolic and discrete~\cite{endy05}. Conversely, quantitative engineering and optimisation of such pathways requires numerical modelling systems that can handle uncertainty and time-series data, e.g., dynamic systems and Ordinary Differential Equations. These are sub-symbolic and only have a continuous parameter space. Hence, to build an automated DBTL cycle driven by AI, the integration of symbolic and numerical calculus is required. 

Nevertheless, most current AI research  is restricted to one side or the other. The data-driven AI, which is represented by deep learning and statistical learning, aims at learning numerical functions from massive annotated data or a simulated environment that allows for a huge number of trial-and-errors. The knowledge-driven AI, such as Inductive Logic Programming and constraint-based problem solving, alternatively focuses on the combinatorial search for a symbolic hypothesis in a discrete space while using pre-defined domain knowledge to prune the search space. Several frameworks have been proposed to combine the two AI paradigms, such as Statistical Relational AI and Neuro-Symbolic Learning. However, to build a gradient-based end-to-end learning pipeline, most of these approaches try to replace symbolic reasoning with statistical inferences or neural network modules and sacrifice the ability to learn relational hypotheses and exploit symbolic domain knowledge, which is critical in synthetic biology.

\begin{figure}[t!]
  \centering
  \includegraphics[width=1\columnwidth]{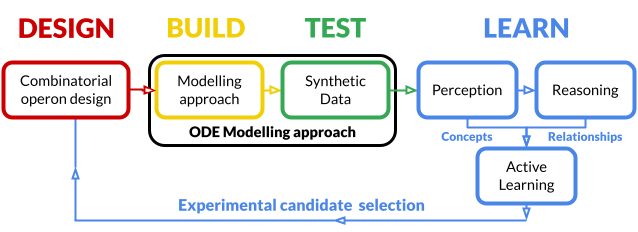}
  \caption{{\small Integrating numerical and symbolic modelling for automated bioengineering\label{fig1}}}
  \vspace{-1.5em}
\end{figure}

In this paper, we proposed to employ the newly proposed Abductive Meta-Interpretive Learning ($Meta_{Abd}$) to improve the DBTL cycle for automated biodesign engineering, which is illustrated by Fig.~\ref{fig1}. $Meta_{Abd}$ is a novel machine learning approach that combines abduction and induction to enable symbolic knowledge induction from numerical data. Specifically, it is designed for simultaneously optimising symbolic hypotheses such as differential equation structures or genetic constructs based on biological background knowledge and fitting the unknown parameters in the models via numerical optimisation. 

We aim to verify the effectiveness of our proposed approach with synthetic time series data from fluorescent protein production, in which the specific genetic  design provides the discrete control variables to be optimised.

\section{Related Work}
\vspace{-.5em}
Computational scientific discovery has been a major research topic since the early days of Artificial Intelligence. \citeauthor{King04:RobotScientist}~(\citeyear{King04:RobotScientist}) learns genetic function by utilising abductive logical reasoning in inductive logic programming, which is a general framework for logic-based machine learning. \citeauthor{Dzeroski95:Lagrange}~(\citeyear{Dzeroski95:Lagrange}) learn the differential equation structures through a combinatorial search constrained by logical background knowledge and a set of qualitative conditions inferred from the training data. Yet it is challenging for these symbolic-based methods to learn models directly from noisy numeric data and make accurate quantitative predictions. The explosion of computing power in the last decades has paved the way for scientific discovery with evolutionary algorithms and deep learning~\cite{schmidt2009distilling,Udrescueaay2631}. However, due to the lack of expressive power, these models cannot make use of generic biological background knowledge that is represented by formal languages. 

\vspace{-.5em}
\section{The Approach}

$Meta_{Abd}$ is a novel Inductive Logic Programming approach aiming to bridge symbolic and sub-symbolic machine learning in a mutually beneficial way~\cite{dai20:metaabd}. Specifically, $Meta_{Abd}$ takes numerical models, e.g., neural networks and differential systems, as logic predicates for processing noisy raw data, and then uses logical abduction and induction to train numerical models and learn symbolic hypotheses jointly. For example, in synthetic biology the symbolic hypotheses may represent the biological system at multiple levels: describing the transgenic constructs, the proteins expressed from these transgenic constructs, the enzymatic pathways these proteins comprise, and host and environmental interactions with the transgenic system. The behaviour of such an engineered system can be represented by differential equation structures; assembled from simple numerical models describing the interactions cellular molecular species using mass action and 
Michaelis–Menten kinetics. Unlike the previous work, $Meta_{Abd}$ can incorporate complex biological background knowledge and learn numerical model for making quantitative predictions jointly. 

Another component of our approach is Active Learning (AL), which can significantly reduce the cost of learning by letting the machine actively choose the experiments to be conducted~\cite{King04:RobotScientist}. For synthetic biology tasks that usually have large design spaces, this could be one of the most important features for the automated DBTL cycle. 

\vspace{-.5em}
\section{Empirical Study on Synthetic Data}
To develop the $Meta_{Abd}$ framework for synthetic biology, 
we have created a synthetic dataset from a modified version of the Wei{\ss}e model~\cite{WeisseE1038} for an operon of three fluorescent proteins to allow for rapid \emph{in silico} exploration of the feature space. It has been adapted to represent multiple discrete features (e.g. promoter, gene orders, RBS strengths, etc.) mathematically and produces data that reflects biologically coherent emergent behaviour in time.



This modified Wei{\ss}e model is utilised to encode and process experimental data into conceptual system knowledge, which forms the ``Perception module'' aspect of the framework (see Fig.~\ref{fig1}). This encoded knowledge can be symbolically represented and passed to $Meta_{Abd}$ for symbolic regression analysis, forming the ``Reasoning module''. An Active Learning strategy is then employed to generate subsequent experimental candidates through logical abduction. 

\vspace{-.5em}
\section{Future Work}
The synthetic dataset has been deliberately constructed around a biodesign problem that can be addressed experimentally. The DNA-BOT platform enables an automated BASIC DNA assembly framework providing a methodology for the flexible modular assembly of operons in terms of promoter, gene order and RBS strength~\cite{marko20}. The synthetic design will be constructed and analysed with $Meta_{Abd}$ to demonstrate its applicability to a real-world biodesign space.

Further development of this approach will focus on the optimisation of a biosynthetic pathway. The platform will expand beyond gene expression of a multigene operon to incorporate the kinetics of the resulting enzymatic pathway. Active learning will be applied to selectively generate novel hypotheses and propose experiments to enable the inclusion of system design rules from complex enzymatic behaviours while reducing the experimental costs.

\vspace{-.5em}
\small
\bibliographystyle{aaai}
\bibliography{bio_meta_abd}

\end{document}